\documentclass[10pt,twocolumn,letterpaper]{article}

\usepackage{cvpr}
\usepackage{epsfig}
\usepackage{graphicx}
\usepackage{amsmath}
\usepackage{amsfonts}
\usepackage{amssymb}
\usepackage{tabularx}
\usepackage{boldline}
\usepackage{color,soul}
\usepackage{bm}
\usepackage{mathrsfs}

\newcolumntype{M}[1]{>{\centering\arraybackslash}m{#1}}

\def\cvprPaperID{4550} 

\cvprfinalcopy
\ifcvprfinal\fi

\begin{document}

\title{Nesti-Net: Normal Estimation for Unstructured 3D Point Clouds using Convolutional Neural Networks}

\author{
Yizhak Ben-Shabat \\
Mechanical Engineering\\
Techion IIT \\
Haifa, Israel \\
{\tt\small sitzikbs@gmail.com}
\and
Michael Lindenbaum \\
Computer Science\\
Techion IIT \\
Haifa, Israel \\
{\tt\small mic@cs.technion.ac.il}
\and
Anath Fischer\\
Mechanical Engineering\\
Techion IIT \\
Haifa, Israel \\
{\tt\small meranath@technion.ac.il}
}

\date{}

\maketitle
\begin{abstract}
In this paper, we propose a normal estimation method for unstructured 3D point clouds. This method, called Nesti-Net, builds on a new local point cloud representation  which consists of multi-scale point statistics (MuPS), estimated on a local coarse Gaussian grid. This representation is a suitable input to a  CNN architecture. The normals are estimated using a mixture-of-experts (MoE) architecture, which relies on a data-driven approach for selecting the optimal scale around each point and  encourages sub-network specialization. Interesting insights into the network's resource distribution are provided. The scale prediction significantly improves robustness to different noise levels, point density variations and different levels of detail.
We achieve state-of-the-art results on a benchmark synthetic dataset and present qualitative results on real scanned scenes. 
 
\end{abstract}

\section{Introduction}
\label{Sec:intro}

Commodity 3D sensors are rapidly becoming an integral part of autonomous systems. These sensors, \eg RGB-D cameras or LiDAR, provide a 3D point cloud representing the geometry of the scanned objects and surroundings. This raw representation is challenging to process since it lacks connectivity information or structure, and is often incomplete, noisy and contains point density variations. 
In particular, processing it by means of the highly effective convolutional neural networks (CNNs) is problematic because CNNs require structured, grid-like data as input. 

When available,  additional local geometric information, such as the surface normals at each point, induces a partial local structure and improves performance of different tasks such as over-segmentation \cite{ben2018graph}, classification \cite{qi2017pointnet++} and surface reconstruction \cite{guerrero2018pcpnet}.

Estimating the normals from a raw, point-only, cloud, is a challenging task due to difficulties associated with sampling density, noise, outliers, and detail level. The common approach is to specify a local neighborhood around a point  and  to fit a local basic geometric surface (\eg a plane) to the points in this neighborhood. Then the normal is estimated from the fitted geometric entity. The chosen size (or scale) of the neighborhood introduces an unavoidable tradeoff between robustness to noise and accuracy of fine details. A large-scale neighborhood over-smoothes sharp corners and small details but is otherwise robust to noise. A small neighborhood, on the other hand, may reproduce the normals more accurately around small details but is more sensitive to noise. Thus it seems that an adaptive, data-driven scale may improve estimation performance.  

We propose a normal estimation method for unstructured 3D point clouds.  It features a mixture-of-experts network for scale prediction, which significantly increases its robustness to different noise levels, outliers, and varying levels of detail. In addition, this method overcomes the challenge of feeding point clouds into CNNs by extending the recently proposed 3D modified Fischer Vector (3DmFV) representation \cite{ben20183dmfv} to encode local geometry on a coarse multi-scale grid. It outperforms state-of-the-art methods for normal vector estimation.
\\
The main contributions of this paper are:
\begin{itemize}
\item A new normal estimation method for unstructured 3D point clouds based on mixture of experts and scale prediction. 
\item A local point representation which can be used as input to a CNN.
\end{itemize}

\begin{figure*}[h]
\centering
	\includegraphics[width=0.98\linewidth]{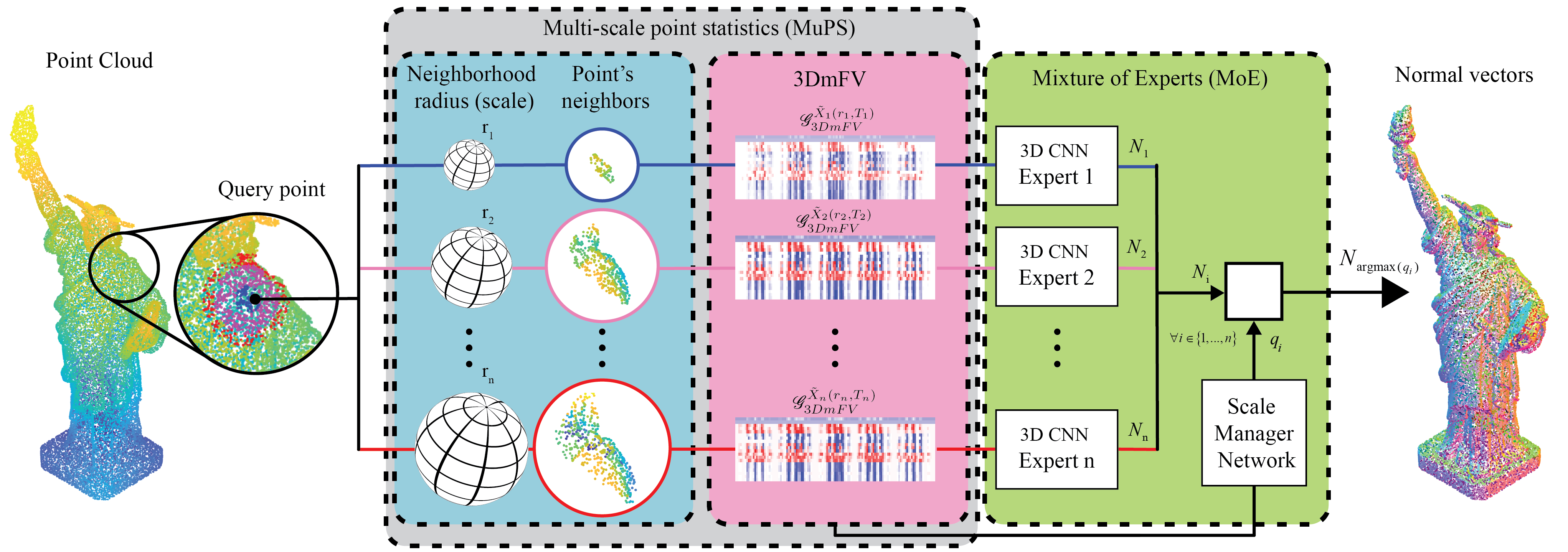}
	\caption{Nesti-Net pipeline for normal estimation. For each point in a given point cloud, we compute a multi-scale point statistics representation (MuPS). Then, a scale manager network is used to determine the optimal scale and uses the corresponding expert sub-network to estimate the normal.}
	\label{fig:approach} 
\end{figure*}
\section{Related-work}
\label{Sec:related-work}

\subsection{Deep learning for unstructured 3D point clouds}
\label{SubSec:rel_work:DL_3D}

The point cloud representation is challenging for deep learning methods because it is both unstructured and point-wise unordered. In addition, the number of points in the point cloud is usually not constant. Several methods were proposed to overcome these challenges. Voxel-based methods embed the point cloud into a voxel grid but suffer from several accuracy-complexity tradeoffs \cite{maturana2015voxnet}. The PointNet approach \cite{Qi_2017_CVPR,qi2017pointnet++} applies a symmetric, order-insensitive, function on a high-dimensional representation of individual points. The Kd-Network \cite{klokov2017escape} imposes a kd-tree structure on the points and uses it to learn shared weights for nodes in the tree. The recently proposed 3DmFV \cite{ben20183dmfv} represents the points by their deviation from a Gaussian Mixture Model (GMM) whose Gaussians are uniformly positioned on a coarse grid. 

In this paper, we propose a point-wise and multi-scale variation of 3DmFV. Instead of generating a structured representation for the entire point cloud, we represent each point and its neighbors within several scales.  

\subsection{Normal estimation}
\label{SubSec:rel_work:Normal_est}

A classic method for normal estimation uses Principal Component Analysis (PCA) \cite{hoppe1992surface}. It first specifies the neighbors within some scale, and then uses PCA regression to estimate a tangent plane. Variants fitting local  spherical surfaces  \cite{guennebaud2007algebraic} or jets \cite{cazals2005estimating} (truncated Taylor expansion) were also proposed. To be robust to noise, these methods usually choose a large-scale neighborhood, leading them to smooth sharp features and to fail in estimating normals near edges. Computing the optimal neighborhood size can decrease the estimation error \cite{mitra2003estimating} but requires the (usually unknown) noise standard deviation value and a costly iterative process to estimate the local curvature and additional density parameters. 

Other approaches rely on using Voronoi cells of point clouds \cite{amenta1999surface, merigot2011voronoi, dey2006provable}. These methods are characterized by robustness to sharp features but are sensitive to noise. To overcome this challenge, Alliez et al. \cite{alliez2007voronoi} proposed PCA-Voronoi approach to create cell sets which group adjacent cells to provide some control over smoothness. While many of these methods hold theoretical guarantees on approximation and robustness, in practice they rely on a preprocessing stage in the presence of strong or unstructured noise in addition to a fine-tuned set of parameters.   

A few deep learning approaches have been proposed to estimate normal vectors from unstructured point clouds. Boulch et al. proposed to transform local point cloud patches into a 2D Hough space accumulator by randomly selecting point triplets and voting for that plane's normal. Then, the normal is estimated from the accumulator by designing explicit criteria \cite{boulch2012fast} for bin selection or, more recently, by training a 2D CNN \cite{boulch2016deep} to estimate it continuously as a regression problem. This method does not fully utilize the 3D information since it loses information during the transformation stage. We reffer to this method as HoughCNN in the evaluation section. A more recent method, PCPNnet \cite{guerrero2018pcpnet}, uses a PointNet \cite{Qi_2017_CVPR} architecture on local point neighborhoods of multiple scales. It achieves good normal estimation performance and has been extended to estimating other surface properties. However, it processes the multi-scale point clouds jointly and does not select an optimal scale. This type of architecture tends to encourage averaging during training rather than specialization \cite{jacobs1991adaptive}.     

In this paper we propose a method that approximates the local normal vector using a point-wise, multi-scale 3DmFV representation which serves as an input to a deep 3D CNN architecture. In addition, we learn the neighborhood size that minimizes the normal estimation error using a mixture of experts (MoE) \cite{jacobs1991adaptive}, which encourages specialization.  

\subsection{Representing point clouds using 3DmFV}
\label{SubSec:rel_work:fisher_vecs_pc}
The 3DmFV representation for point clouds \cite{ben20183dmfv} achieved good results for point cloud classification using a 3D CNN.  See Section \ref{subsec:approach:pw-ms-3dmfv} for details. In this paper we propose the Multi-scale Point Statistics (MuPS) representation, which extends the 3DmFV and computes a point-wise multi-scale 3DmFV.
\section{Approach}
\label{Sec:approach}
The proposed method is outlined in Figure \ref{fig:approach}. It receives a 3D point cloud as input and consists of two main stages. In the first stage, we compute a multiscale point representation, denoted MuPS. In the second stage we feed it into a mixture-of-experts (MoE) CNN architecture and estimate the normal at each point as output. The stages are detailed below.

\subsection{MuPS - Multi-scale point statistics}
\label{subsec:approach:pw-ms-3dmfv}
MuPS is a local multi-scale representation which computes point statistics on a coarse Gaussian grid. It builds on the well-known Fisher Vector (FV) \cite{sanchez2013image}, and the recently proposed 3DmFV representation \cite{ben20183dmfv}. Therefore, we first outline the FV and the 3DmFV, and then continue to the  MuPS representation and its attractive properties. 
\\
\textbf{FV and 3DmFV for 3D point clouds}: Given a set of $T$ 3D points $X=\{ \bm{p}_t \in \mathbb{R}^3, t=1,...T\}$ and a set of parameters for a $K$ component GMM, $\lambda=\{(w_k,\mu_k,\Sigma_k), k=1,...K\}$, where $w_k,\mu_k,\Sigma_k$ are the mixture weight, center, and covariance matrix of the \mbox{$k$-th} Gaussian. The likelihood of a single 3D point  $\bm{p}$ associated with the $k$-th Gaussian density is 
    \begin{equation}
    	u_k(\bm{p}) = \frac{1}{(2\pi)^{D/2}|\Sigma_k|^{1/2}}\exp\left\{-\frac{1}{2}(\bm{p}-\mu_k)'\Sigma_k^{-1}(\bm{p}-\mu_k)\right\}.
    \end{equation}
\\
Therefore, the likelihood of a single point associated with the GMM density is:
    \begin{equation}
    	u_\lambda(\bm{p}) = \sum_{k=1}^{K}w_ku_k(\bm{p}). 
    \end{equation}
\\ 
The 3DmFV uses a uniform GMM on a coarse ${m\times m\times m}$ 3D grid, where $m$ is an adjustable parameter usually chosen to be from $m = 3$ to $8$. The weights are set to be $w_k = \frac{1}{K}$, the standard deviation is set to be $\sigma_k = \frac{1}{m}$, and the covariance matrix to be $ \Sigma_k = \sigma_kI $. Although the parameters in GMMs are usually set using maximum likelihood estimation, here uniformity is crucial for shared weight filtering (convolutions). 

The FV is expressed as the sum of normalized gradients for each point $p_t$. The 3DmFV is specified similarly using additional symmetric functions, \ie $min$ and $max$. They are symmetric in the sense proposed in \cite{Qi_2017_CVPR} and are therefore adequate for representing the structureless and orderless set of points. Adding these functions makes the representation more informative and the associated classification more accurate \cite{ben20183dmfv}: 
    \begin{equation}
    \label{eq:Fisher}
    	\mathscr{G}_{FV_\lambda}^X = \sum_{t=1}^TL_\lambda\nabla_\lambda\log u_\lambda(\bm{p}_t),
    \end{equation}
        \begin{equation} \label{eq:3DmFV_definition}
	\mathscr{G}_{3DmFV_\lambda}^X = \left[ \begin{array}{c} 
	\left. \sum_{t=1}^TL_\lambda\nabla_\lambda\log u_\lambda(\bm{p}_t) \right|_{\lambda=\alpha,\mu,\sigma} \\
	 \left. \max_t(L_\lambda\nabla_\lambda\log u_\lambda(\bm{p}_t)\right|_{\lambda=\alpha,\mu,\sigma} \\
	 \left. \min_t(L_\lambda\nabla_\lambda\log u_\lambda(\bm{p}_t))\right|_{\lambda=\mu,\sigma} \end{array}\right]
	\end{equation}
where $L_\lambda$ is the square root of the inverse Fisher Information Matrix, and the normalized gradients are: 
	 \begin{equation} \label{eq:dev_w}
	 	\mathscr{G}_{\alpha_k}^X = \frac{1}{\sqrt{w_k}} \sum_{t=1}^T(\gamma_t(k)-w_k),
	 \end{equation}
	 \begin{equation} \label{eq:dev_mu}
	 	\mathscr{G}_{\mu_k}^X = \frac{1}{\sqrt{w_k}} \sum_{t=1}^T \gamma_t(k) \left( \frac{\bm{p}_t-\mu_k}{\sigma_k} \right),
	 \end{equation}
	 \begin{equation} \label{eq:dev_sig}
	 	\mathscr{G}_{\sigma_k}^X = \frac{1}{\sqrt{2w_k}} \sum_{t=1}^T \gamma_t(k) \left[ \frac{(\bm{p}_t-\mu_k)^2}{\sigma_k^2}-1 \right].
	 \end{equation}    
Here, we follow \cite{krapac2011modeling} and ensure that $u_\lambda(x)$ is a valid distribution by changing the variable $w_k$ to $\alpha_k$ to simplify the gradient calculation using : 
    \begin{equation}\label{eq:wk}
    	w_k = \frac{exp(\alpha_k)}{\sum_{j=1}^{K}exp(\alpha_j)}.
    \end{equation}
\\
In addition, $\gamma_t(k)$ expresses the soft assignment of point $\bm{p}_t$ to Gaussian $k$, as obtained from the derivatives: 
\begin{equation}\label{eq:gamma}
	\gamma_t(k) = \frac{w_k u_k(\bm{p}_t)}{\sum_{j=1}^{K} w_j u_j(\bm{p}_t)}.
\end{equation}  
\\
The FV and 3DmFV are normalized by the number of points in order to be sample-size independent \cite{sanchez2013image}:
\begin{equation} \label{eq:FV_norm_T}
\mathscr{G}_{FV_\lambda}^X \leftarrow \frac{1}{T}\mathscr{G}_{FV_\lambda}^X,
\mathscr{G}_{3DmFV_\lambda}^X \leftarrow \frac{1}{T}\mathscr{G}_{3DmFV_\lambda}^X.
\end{equation} 
\\
Note that these are global representations which are applied to the entire set, \ie, the entire point cloud. 
\\
\textbf{MuPS definition} : For each point $\bm{p}$ in point set $X$ we first extract $n$ point subsets $\tilde{X}_i(r_i)|_{i=1,...n} \subset X $ which contain $T_i(\bm{p},r_i)$ points and lie within a distance of radius $r_i$ from $\bm{p}$. We refer to each of these subsets as a scale. Note that each scale may contain a different number of points.  For scales with many points, we set a maximal point threshold, and sample a random subset of $T_{max}$ points for that scale. Here, $r_i$ and $T_{max}$ are design parameters.
Next, the scales (subsets) are independently translated and uniformly scaled so that they fit into a zero-centered unit sphere with $\bm{p}$ mapped to the origin. Then, the 3DmFV representation is computed for each scale relative to a  Gaussian grid positioned around the origin; see above. Concatenating the 3DmFVs of all scales yields the MuPS representation: 
\\
\begin{equation}
\label{eq:mups}
	\mathscr{G}_{MuPS}^{\bm{p}} = \left(   
	\mathscr{G}_{3DmFV}^{\tilde{X}_1(r_1)},...,\mathscr{G}_{3DmFV}^{\tilde{X}_n(r_n)} \right).
\end{equation}
\\
\textit{MuPS properties}: The MuPS representation overcomes the main challenges associated with feeding point clouds into CNNs. The symmetric functions make it independent of the number and order of points within each scale. In addition, the GMM gives it its grid structure, necessary for the use of CNNs. Furthermore, the multi-scale representation incorporates description of fine detail as well as robustness to noise. 

\subsection{The Nesti-Net architecture}
\begin{figure}
    \centering
    \includegraphics{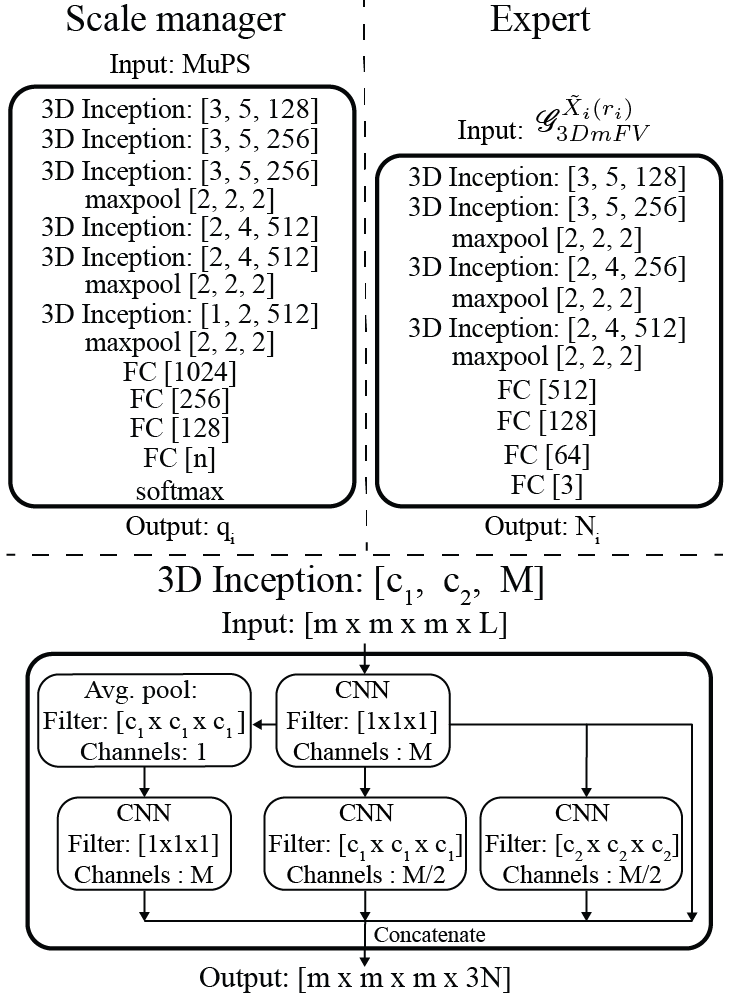}
    \caption{The mixture of experts and 3D Inception module architecture details. The scale manager and experts use several convolutional and maxpooling layers followed by fully connected layers.}
    \label{fig:approach:specific_architectture}
\end{figure}

The deep network architecture is outlined in Figure \ref{fig:approach} (the green part). It is a mixture-of-experts architecture \cite{jacobs1991adaptive} which consists of two modules: a scale manager network module, and an experts module. The MoE architecture was chosen in order to overcome the averaging effect of typical networks when solving a regression problem. 
\\
\textit{Scale manager network}: This module receives the MuPS representation as input and processes it using several 3D Inception inspired convolutions, and maxpool layers, followed by four fully connected layers, after which a softmax operator is applied.  The architecture is specified in the top left part of Figure \ref{fig:approach:specific_architectture}.   The output is a vector of $n$ scalars $q_i$, which can be intuitively interpreted as the probability of expert $i$ to estimate the normal correctly.
\\
\textit{Experts}: The normal is estimated using $n$ separate "expert" networks. Each is a multi-layered 3D Inception inspired CNN followed by four  fully connected layers. The MuPS representation is distributed to the experts. This distribution is a design choice. We obtained the best results when feeding each scale to two different experts in addition to one expert which receives the entire MuPS representation as input. Specifically,  Nesti-Net uses 7 experts: experts 1-2 receive the smallest scale (1\%), 3-4 the medium scale (3\%), 5-6 the large scale (5\%), and expert 7 receives all the scales. The last layer of each expert outputs a three-element vector $N_i=(N_x, N_y, N_z)_i$. The final predicted normal (for point $\bm{p}$) is $N_{argmax(q_i)}$, \ie, the normal associated with the expert expected to give the best results. The architecture is specified in the top right of Figure \ref{fig:approach:specific_architectture}.   
\\
\textit{Loss function}: 
We train the network to minimize the difference  between a predicted normal $N_{i}$ and a ground truth normal $N_{GT}$. This  difference is quantified by the metric $D_N=\sin{\theta}$, where the angle $\theta$ is the difference between the vectors, and $D_N$ is calculated as the magnitude of the cross product between these two vectors; see Eq. \ref{eq:loss}. In addition, to encourage specialization of each expert network, we follow \cite{jacobs1991adaptive} and minimize the loss:
\begin{equation}
\label{eq:loss}
    L = \sum_{i=1}^{n}q_i \cdot D_N =  \sum_{i=1}^{n}q_i \frac{\left\lVert N_{i} \times N_{GT}\right\rVert}{\left\lVert N_{i}\right\rVert \cdot \left\lVert N_{GT} \right\rVert}.
\end{equation}
Using this loss, each expert is rewarded for specializing in a specific input type. Note that during training, all $n$ normal vectors are predicted and used to compute the loss and derivatives. However, at test time, we compute only one normal, which is associated with the maximal $q_i$. 
\section{Evaluation}
\label{Sec:results}

\subsection{Datasets} 
For training and testing we used the PCPNet shape dataset \cite{guerrero2018pcpnet}. The trainset consists of 8 shapes: four CAD objects (fandisk, boxunion, flower, cup) and four high quality scans of figurines (bunny, armadillo, dragon and turtle). All shapes are given as triangle meshes and densely sampled with 100k points. The data is augmented by introducing Gaussian noise for each point's spacial location with a standard deviation of  0.012, 0.006, 0.00125 w.r.t the bounding box. This yields a set with 3.2M points and 3.2M corresponding training examples. The test set consists of 22 shapes, including figurines, CAD objects, and analytic shapes. For evaluation we use the same 5000 point subset per shape as in \cite{guerrero2018pcpnet}.

For qualitative testing on scanned data, we used the NYU Depth V2  dataset  \cite{SilbermanECCV12} and the recent ScanNet dataset \cite{dai2017scannet}, which include RGB-D images of indoor scenes.  

\subsection{Training details}
\label{SubSec:results:trainnig_details}
All variations of our method were trained using 32,768 (1024 samples$\times32$ shapes) random subsets of the 3.2M training samples at each epoch. For each point, we extract 512 neighboring points enclosed within a sphere of radius $r$. For neighborhoods with more than 512 points, we perform random sampling, and for those with fewer points we use the maximum number of points available. For the MuPS representaiton we chose to use an $m=8$ Gaussian grid. We used Tensorflow on a single NVIDIA Titan Xp GPU.   

\subsection{Normal estimation performance}
\label{SubSec:results:baseline_n_est}
 We use the RMS normal estimation error metric for comparing the proposed NestiNet to other deep learning based  \cite{guerrero2018pcpnet, boulch2016deep} and geometric methods \cite{hoppe1992surface, cazals2005estimating}. 
 We also analyze robustness for two types of data corruptions (augmentations): 
\begin{itemize}
    \item Gaussian noise - perturbing the points with three levels of noise specified by $\sigma$, given as a percentage of the bounding box.
    \item Density variation - selecting a subset of the points based on two sampling regimes: gradient, simulating effects of distance from the sensor, and stripes, simulating occlusions.    
\end{itemize}
For the geometric methods, we show results for three different scales: small, medium and large, which correspond to 18, 112, 450 nearest neighbors. For the deep learning based methods we show the results for the single-scale (ss) and multi-scale (ms) versions.  Additional evaluation results using other metrics are available in the supplemental material. 

Table \ref{table:results:baselines} shows the unoriented normal estimation results for the methods detailed above. It can be seen that our method outperforms all other methods across all noise levels and most density variations.  It also shows that both PCA and Jet perform well for specific noise-scale pairs. In addition, for PCPNet and HoughCNN, using a multi-scale approach only mildly improves performance. 

    \begin{table*} 
	\centering	
		\tabcolsep = 0.001\textwidth
		\begin{tabular}{| M{0.12\textwidth} | M{0.1\textwidth} | M{0.06\textwidth}| M{0.06\textwidth} | M{0.06\textwidth}|
		M{0.06\textwidth}| M{0.06\textwidth} | M{0.06\textwidth}|  M{0.07\textwidth} | M{0.07\textwidth} | M{0.08\textwidth} | M{0.08\textwidth} |} 
			\hline
			\centering\textbf{Aug.} &\centering\textbf{Our Nesti-Net} & \multicolumn{3}{c|}{\textbf{PCA} \cite{hoppe1992surface} }  &
			\multicolumn{3}{c|}{\textbf{Jet} \cite{cazals2005estimating} } & 	\multicolumn{2}{c|}{\textbf{PCPNet} \cite{guerrero2018pcpnet} } & \multicolumn{2}{c|}{\textbf{HoughCNN} \cite{boulch2016deep}}
 			\tabularnewline
 			\hline
 			scale &&small&med&large&small&med&large&ss&ms&ss&ms\\
 			\hlineB{2}
 		    None               &\textbf{6.99}&8.31&12.29&16.77&7.60&12.35&17.35&9.68&9.62&10.23&10.02\\
 		    \hline
 			\textbf{Noise}     &&&&&&&&&&& \\
 			$\sigma = 0.00125$ &\textbf{10.11}&12.00&12.87&16.87&12.36&12.84&17.42&11.46&11.37&11.62&11.51\\
 			$\sigma = 0.006$   &\textbf{17.63}&40.36&18.38&18.94&41.39&18.33&18.85&18.26&18.87&22.66&23.36\\
 			$\sigma = 0.012$   &\textbf{22.28}&52.63&27.5 &23.5 &53.21&27.68&23.41&22.8&23.28&33.39&36.7\\
		    \hline
		    \textbf{Density}   &&&&&&&&&&& \\
		    Gradient           &9.00&9.14&12.81&17.26&\textbf{8.49}&13.13&17.8&13.42&11.7 &11.02&10.67\\
		    Stripes            &\textbf{8.47}&9.42&13.66&19.87&8.61&13.39&19.29&11.74&11.16&12.47&11.95\\
		    \hline
		    \textbf{average}   &\textbf{12.41}&21.97&16.25&18.87&21.95&16.29&19.02&14.56&14.34&16.9&17.37 \\
		    \hline
		\end{tabular}
	\caption{Comparison of the RMS angle error for unoriented normal vector estimation of our Nesti-Net method to classic geometric methods (PCA \cite{hoppe1992surface} , Jet \cite{cazals2005estimating}) with three scales, and deep learning methods (PCPNet \cite{guerrero2018pcpnet}, HoughCNN \cite{boulch2016deep})}
	\label{table:results:baselines}
\end{table*}

\begin{figure}
\centering
	\includegraphics[width=0.98\linewidth]{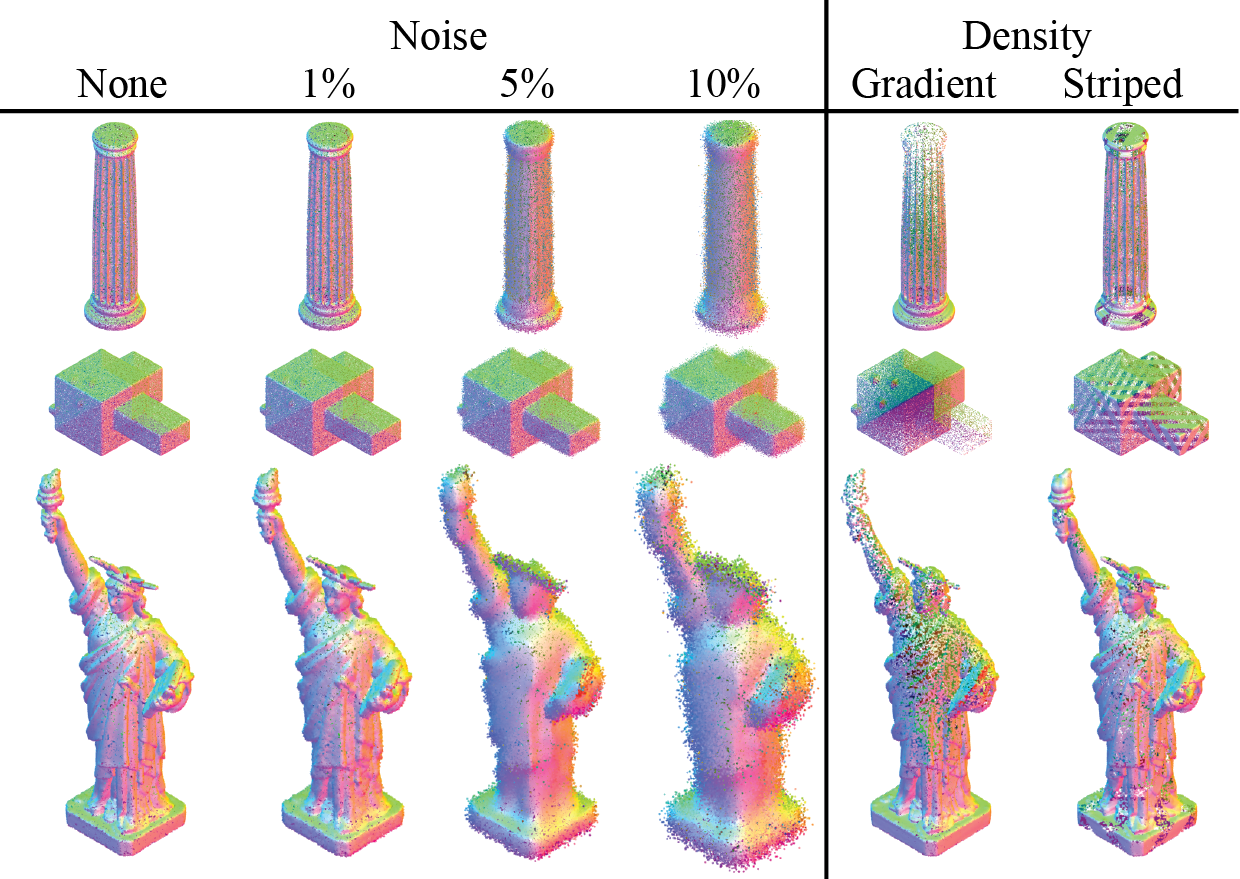}
	\caption{Nesti-Net normal prediction results for different noise levels (columns 1-4), and density distortions (columns 5-6). The colors of the points are normal vectors mapped to RGB. This figure is best viewed digitally on a large screen.}
	\label{fig:results_normals_visualiztion}
\end{figure}

\begin{figure}
\centering
	\includegraphics[width=0.98\linewidth]{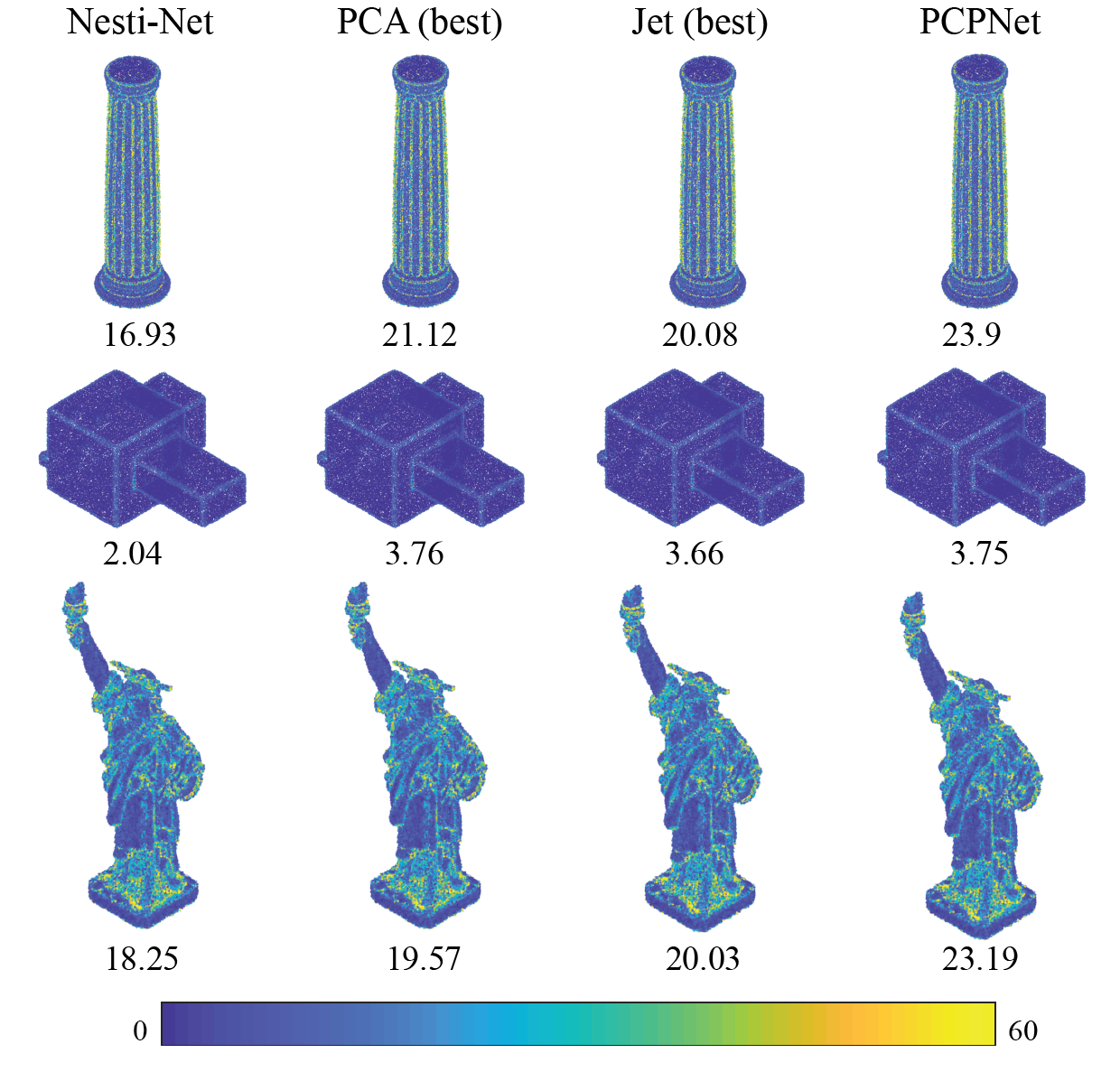}
	\caption{Normal estimation error results for Nesti-Net compared to other methods for three types of point clouds with low noise level ($\sigma = 1\%$). The colors of the points correspond to angular difference, mapped to a heatmap ranging from 0-60 degrees; see bottom color bar. The number under each point cloud is its RMS error.}
	\label{fig:results_error_comparison} 
\end{figure}

\begin{figure}
\centering
	\includegraphics[width=0.98\linewidth]{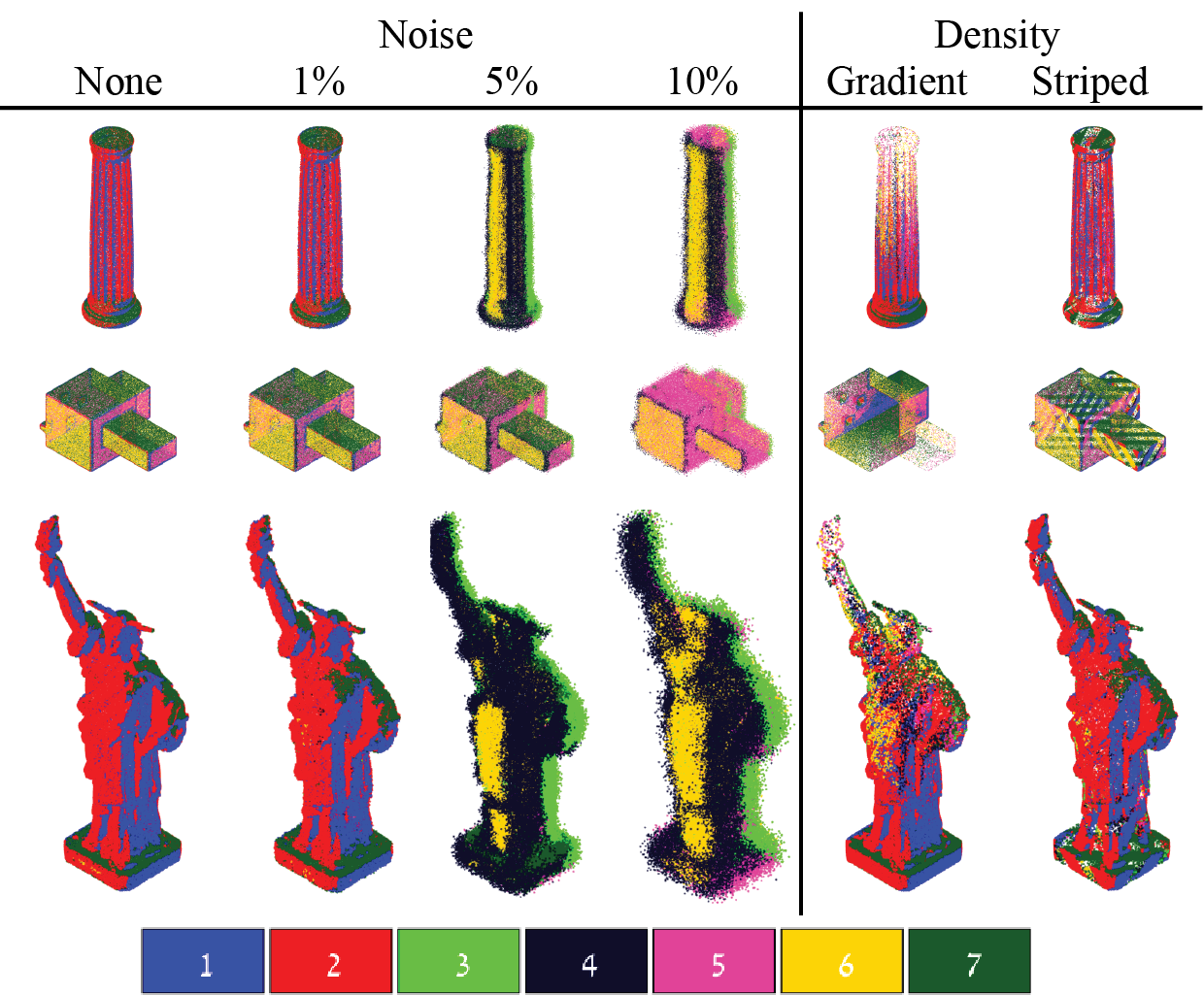}
	\caption{Nesti-Net predicted experts (scales). Each color represents the predicted expert for optimal normal estimation. Color coding is given at the bottom.}
	\label{fig:results_expert_visualiztion} 
\end{figure}

\begin{figure}
\centering
\begin{tabular}{cc}
     \includegraphics[width=0.22\textwidth]{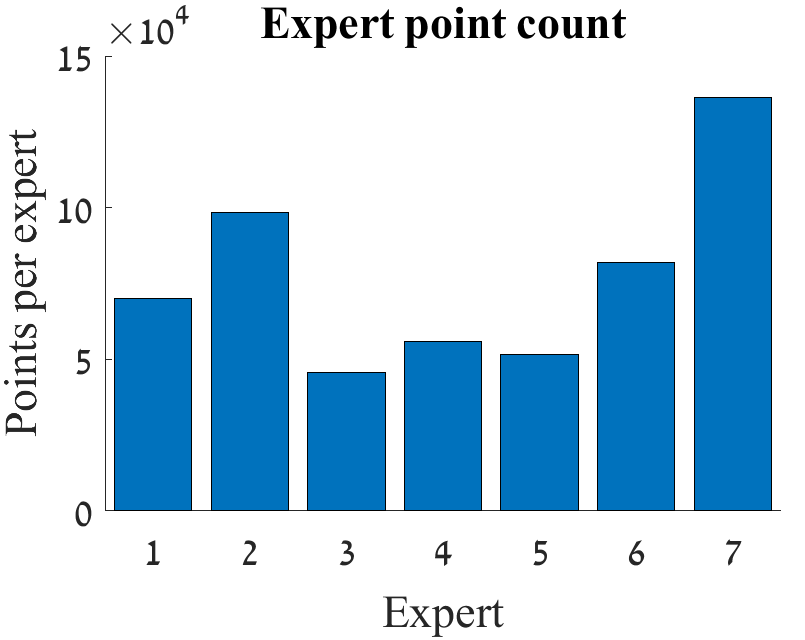}
     &
	\includegraphics[width=0.22\textwidth]{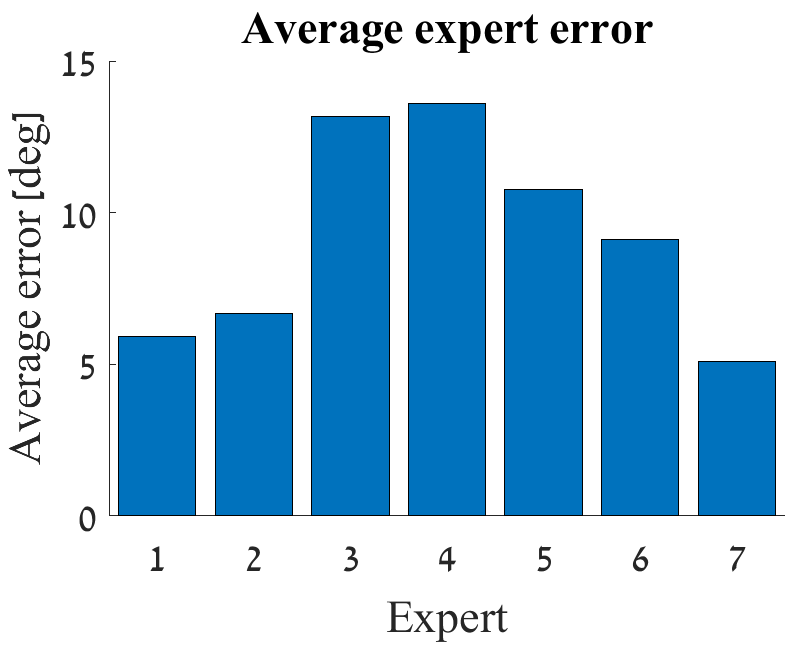}
	\end{tabular}
	\caption{Nesti-Net expert (scale) prediction statistics. Number of points assigned to each expert (left), and average expert error (right).}
	\label{fig:results_expert_stats} 
\end{figure}

Figure \ref{fig:results_normals_visualiztion}  illustrates Nesti-Net's results on three point clouds. For visualization, the normal vector is mapped to the RGB cube. It shows that for complex shapes (pillar, liberty) with high noise levels, the  general direction of the normal vector is predicted correctly, but, the fine details and exact normal vector are not obtained. For a basic shape (Boxysmooth) the added noise does not affect the results substantially. Most notably, Nesti-Net shows robustness to point density corruptions. 
The angular error in each point is visualized in Figure \ref{fig:results_error_comparison} for the different methods using a heat map. For PCA and Jet we display the best result out of the three scales (small, medium, and large, specified above), and for PCPNet the best out of its single-scale and multi-scale options. For all methods, it can be seen that more errors occur near edges, corners and small regions with a lot of detail and high curvature. Nesti-Net suffers the least from this effect due to its scale manager, which allows it to adapt to the different local geometry types. 
Figure \ref{fig:results_expert_visualiztion} shows the performance of the scale manager network. A color is assigned to each expert and the chosen expert color is visualized over the point cloud. This provides some insight regarding each expert's specialization. For example, the figure shows that experts 1, 2 (small scale) specialize in points in regions with high curvatures (near corners).  Experts 3 and 4 (medium scale) specialize in the complex cases where multiple surfaces are close to each other, or in the presence of noise. As for the large-scale experts, expert 5 specializes in planar surfaces with normal vectors, which have a large component in the $x$ direction, whereas expert 6 specializes in planar surfaces, which have a large component in the $y$ direction. Expert 5 also specializes in very noisy planar surfaces with a large component in the $z$ direction. Expert 7 (combined scales) plays multiple roles; it handles points on planar surfaces which have a large component in the $z$ direction, complex geometry, and low to medium noise.  Figure \ref{fig:results_expert_stats} shows the number of points assigned to each expert for all points in the test set, and the average error per expert. It shows an inverse relation between the number of points assigned to an expert and its average error: the more points assigned to the expert, the lower the error. This is consistent with the definition of the cost function. 
Timing performance and visualization of additional results are provided in the supplemental material.


\subsection{Scale selection performance}
\label{Sec:results:ablation}
We analyze the influence of scale selection on the normal estimation performance. We create several ablations of our method.
\begin{itemize}
    \item ss - A single scale version which directly feeds a 3DmFV representation into a CNN architecture (a single-scale MuPS). 
    \item ms - A multi-scale version which feeds the MuPS representation into a CNN architecture.
    \item ms-sw - A multi scale version which first tries to estimate the noise level and then feeds the 3DmFV representation of the corresponding input scale into different sub-networks for a discrete number of noise levels (switching). Note that for this version, the noise level is provided during training. 
    \item NestiNet - The method described in Section \ref{Sec:approach} which uses an MoE network to learn the scale.
\end{itemize}
Details of the architectures for the above methods are provided in the supplemental material. 

Table \ref{table:results:ablation} summarizes the results of the scale selection performance analysis. It shows that Nesti-Net's scale selection performs better than all other variations. This is due to the trained scale-manager network within the MoE. The single-scale version performs well for specific noise-scale pairs but inferior performance for an inadequate scale selection. The multi-scale variations show improvement; however, selecting the correct scale yields improved performance over concatenating multiple scales. The main advantage of Nesti-Net over the switching variation is that the scale prediction is unsupervised, \ie, does not need  the additional noise parameters as input during training.
    \begin{table} 
	\centering	
		\tabcolsep = 0.001\textwidth
		\begin{tabular}{| M{0.11\textwidth} | M{0.05\textwidth} | M{0.05\textwidth}| M{0.05\textwidth}| M{0.07\textwidth} | M{0.06\textwidth}| M{0.06\textwidth}|} 
			\hline
			\centering\textbf{Aug.} & \multicolumn{2}{c|}{\centering\textbf{ss}} & \textbf{ms} & \textbf{ms-sw} & \multicolumn{2}{c|}{\textbf{NestiNet}}
 			\tabularnewline
 			\hline
 			scale &0.01&0.05&0.01 0.05&0.01 0.05&0.01 0.05&0.01 0.03 0.05 \\ 
 			\hlineB{2}
 		    None &9.32&12.73&10.83&7.88&7.76&\textbf{6.99} \\
 		    \hline
 			\textbf{Noise} &&&&&& \\
 			$\sigma = 0.00125$ &11.31&13.36&12.98&10.46&10.29&\textbf{10.11}\\
 			$\sigma = 0.006$   &36.5&18.37&21.06&18.43&18.45&\textbf{17.63} \\
 			$\sigma = 0.012$  &55.24&23.14&26.03&22.59&22.25&\textbf{22.28}\\
		    \hline
		    \textbf{Density} &&&&&& \\
		    Gradient  &16.61&14.65&12.81&11.89&9.44&\textbf{9.00} \\
		    Stripes   &14.5 &14.57&12.97&10.06&9.65&\textbf{8.47} \\
		    \hline
		    \textbf{average} &23.91&16.14&16.11&13.55&12.97&\textbf{12.41}\\
		    \hline
		\end{tabular}
	\caption{Comparison of the RMS angle error for unoriented normal vector estimation of our method using single-scale (SS), multi-scale (MS), multi-scale with switching (MS-Sw and multi-scale with mixture of experts (Nesti-Net)}
	\label{table:results:ablation}
\end{table}

\subsection{Results on scanned data}
\label{subSec:results:scanned_data}
\begin{figure}
    \centering
    \includegraphics[width=0.98\linewidth]{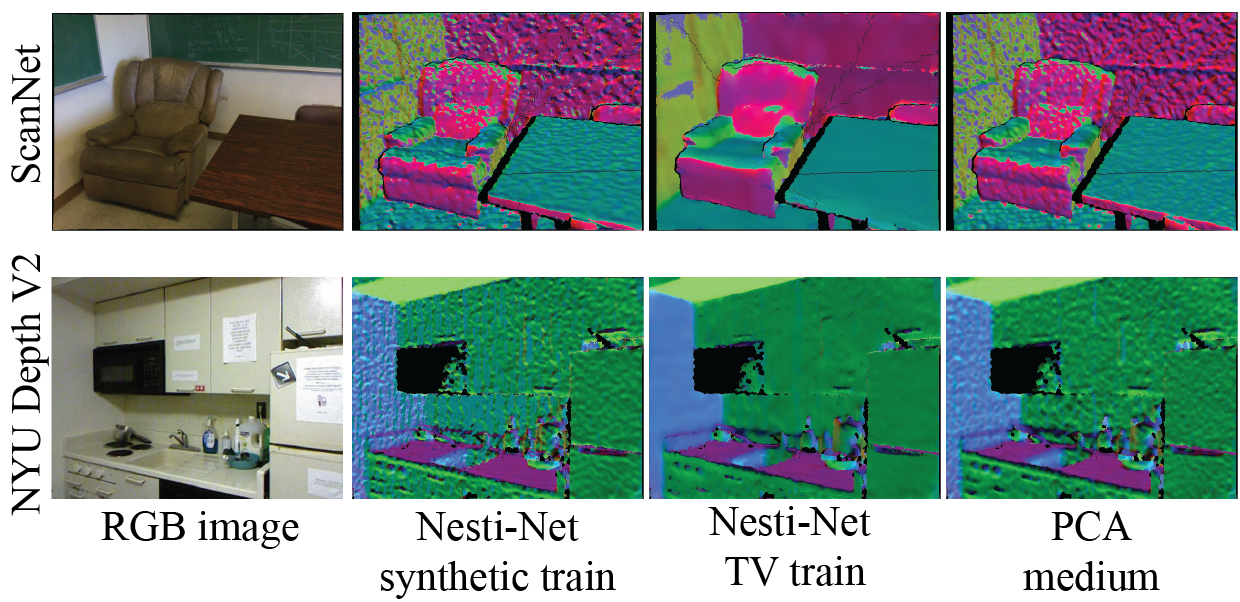}
    \caption{Normal estimation results on scanned point clouds from the ScanNet \cite{dai2017scannet} (top), and NYU Depth V2 dataset \cite{SilbermanECCV12} (bottom). (a) RGB image, (b) PCA results using a medium scale, (c) Nesti-Net results trained on synthetic data (d) Nesti-net results trained on TV algorithm data.}
    \label{fig:results:real_data}
\end{figure}

We show qualitative results on scanned point clouds from the ScanNet \cite{dai2017scannet} and NYU Depth V2 \cite{SilbermanECCV12} datasets in Figure \ref{fig:results:real_data}. For visualization we project the normal vectors' color back to the depth image plane. column (c) shows the results for Nesti-Net, trained on synthetic data with Gaussian noise. The estimated normals reveal the nonsmoothness of the scanned data associated with the  correlated, non-Gaussian, noise signature associated with the scanning process. Essentially it shows normal estimation of the raw data, rather than the desired normal of the underlying surface. The raw point clouds suffer from "bumps" which get bigger as the distance from the sensor increases. 
 Further improvement may be obtained by training Nesti-Net on data corrupted with scanner noise and with ground truth normals, but such data is is currently not available and is difficult to manually label. Instead, we train Nesti-Net with normal vectors obtained from applying a Total Variation (TV) algorithm on the depth map, provided by Ladicky et al. \cite{zeisl2014discriminatively} for the NYU depth V2 dataset. Note that TV substantially smooths fine detail and uses the depth image rather than unstructured point clouds. Column (d) in Figure \ref{fig:results:real_data} shows that after training on the TV data, the normal vector estimation of the underlying surface improves significantly. Column (b) shows the results of PCA with a medium scale for reference, for small radius, the result is significantly noisier and for large radius it over-smooths detail, see supplemental material. Note that Nesti-Net performs the estimation on the raw point cloud and does not use the depth image grid structure. 

\section{Summary}
\label{Sec:summary}

In this work, we propose multi-scale point statistics, a new representation for 3D point clouds that encodes fine and coarse details while using a grid structure. The representation is effective processed by a CNN architecture (Nesti-Net) for provide accurate normal estimation, which can be used for various applications, \eg surface reconstruction. The mixture-of-experts design of the architecture enables the prediction of an optimal local scale and provides insights into the network's resource distribution. 
The proposed representation and architecture achieve state-of-the-art results relative to all other methods and demonstrate robustness to noise and occlusion data corruptions. 

\section{Acknowledegment}
We gratefully acknowledge the support of NVIDIA Corporation with the donation of the Titan Xp GPU used for this research.

{\small
\bibliographystyle{ieee}
\bibliography{references}
}

\twocolumn[\section*{\centering Nesti-Net: Normal Estimation for Unstructured 3D Point Clouds using Convolutional Neural Networks}]
\vspace*{12pt}
\subsection*{Supplementary Material}
\label{Sec:appendix}
\renewcommand{\thesubsection}{\Alph{subsection}}

\subsection{Scale selection methods: architecture details}
In Section \ref{Sec:results:ablation} we report the performance of several ablations of our method. Here we detail the architecture of the following ablations: 
\begin{itemize}
    \item ss - A single scale version which directly feeds a 3DmFV representation into a CNN architecture (a single-scale MuPS); see Table \ref{appendix:table:ablations_1}.
    \item ms - A multi-scale version which feeds the full MuPS representation into a CNN architecture; see Table \ref{appendix:table:ablations_1}. 
    \item ms-sw - A multi scale version which first attempts to estimate the noise level and then feeds the 3DmFV representation of the corresponding input scale into two sub-networks for two noise levels (switching). Note that for this version, the noise level is provided during training and we use a predetermined threshold for the sub-network selection; see Table \ref{appendix:table:ablations_2}.
\end{itemize}

\begin{table}[b]
    \centering
    \begin{tabular}{c|c}
         \textbf{ss} & \textbf{ms}  \\
         \hline
         3D Inception(3,5,128)& 3D Inception(3,5,128) \\
         3D Inception(3,5,256)& 3D Inception(3,5,256)\\
         3D Inception(3,5,256)& 3D Inception(3,5,256) \\
         maxpool(2,2,2)& maxpool(2,2,2)\\
         3D Inception(3,5,512)& 3D Inception(3,4,512) \\
         3D Inception(3,5,512)&  3D Inception(3,4,512)\\
         maxpool(2,2,2)& maxpool(2,2,2) \\
         FC(1024)&FC(1024)\\
         FC(256)&FC(256)\\
         FC(128)&FC(128)\\
         FC(3)&FC(3)\\
    \end{tabular}
    \caption{Ablation architecture details for single-scale (ss) and multi-scale (ms).}
    \label{appendix:table:ablations_1}
\end{table}

\begin{table}[b]
    \centering
    \begin{tabular}{c|c}
    \multicolumn{2}{c}{\textbf{ms-sw}}\\
    \hline
    noise estimation net & normal estimation net \\
    \hline
    3D Inception(3,5,128)& 3D Inception(3,5,128) \\
    3D Inception(3,5,256)& 3D Inception(3,5,256)\\
    3D Inception(3,5,256)& 3D Inception(3,5,256) \\
    maxpool(2,2,2)& maxpool(2,2,2)\\
    3D Inception(3,5,512)& 3D Inception(3,4,512) \\
    3D Inception(3,5,512)&  3D Inception(3,4,512)\\
    maxpool(2,2,2)& maxpool(2,2,2) \\
    FC(1024)&FC(1024)\\
    FC(256)&FC(256)\\
    FC(128)&FC(128)\\
    FC(1)&FC(3)\\
    \end{tabular}
    \caption{Ablation architecture details for multi-scale with switching. First the noise is estimated and then the input is fed into the corresponding scale network according to a threshold. The network for both scales is constructed identically.}
    \label{appendix:table:ablations_2}
\end{table}
\subsection{Normal estimation performance analysis}
\label{subSec:appendix:results_vis}
In Section \ref{SubSec:results:baseline_n_est} we report the RMS error metric results for comparison to other methods. The RMS error favors averaging methods. For example, near corners, it will reward  a method that estimates an average normal direction rather than a method that estimates the normal of the wrong plane. Therefore, a complimentary metric is required to negate this effect. We use the proportion of good points metric (PGP$\alpha$), which computes the percentage of points with an error less than $\alpha$; \eg, PGP10 computes the percentage of points with angular error of less than 10 degrees. Table \ref{appendix:results:pgp10} reports the results of PGP10 and Table \ref{appendix:results:pgp5} the results of PGP5 for the baseline methods compared to Nesti-Net. 

\begin{table*}
    \centering
            \begin{tabular}{|M{0.12\textwidth}|M{0.06\textwidth}|M{0.06\textwidth}|M{0.06\textwidth}|M{0.06\textwidth}|M{0.06\textwidth}|M{0.06\textwidth}|M{0.06\textwidth}|M{0.06\textwidth}|M{0.09\textwidth}|}
    \hline
\textbf{Aug.} & \multicolumn{2}{c|}{\textbf{PCPNet} \cite{guerrero2018pcpnet}}& \multicolumn{3}{c|}{\textbf{Jet} \cite{cazals2005estimating}} & \multicolumn{3}{c|}{\textbf{PCA} \cite{hoppe1992surface}} & \textbf{NestiNet}\\
\hline
Scale&ss&ms&small&med&large&small&med&large&MoE\\
\hline
None &0.8364&0.8404&0.8802&0.7509&0.6584&0.8686&0.7409&0.6606&\textbf{0.9120}\\
\hline
Noise   &&&&&&&&&\\
$\sigma = 0.00125$ &0.8013&0.8031&0.7346&0.7447&0.6575&0.7712&0.7378&0.6598&\textbf{0.8384}\\
$\sigma = 0.006$ &0.6667&0.6294&0.1006&0.6397&0.6311&0.1101&0.6402&0.6301&\textbf{0.7164}\\
$\sigma = 0.01$ &0.5546&0.5124&0.0377&0.3827&0.547&0.04063&0.394&0.5462&\textbf{0.6123}\\
\hline
Density &&&&&&&&&\\
Gradient&0.7801&0.8062&0.8848&0.7695&0.6401&0.8731&0.7624&0.6366&\textbf{0.9003}\\
Striped&0.7967&0.8076&0.8743&0.7504&0.6001&0.8609&0.7379&0.5879&\textbf{0.8929}\\
\hline
Average&0.7393&0.7332&0.5854&0.6730&0.6224&0.5874&0.6689&0.6202&\textbf{0.8120}\\
\hline
    \end{tabular}
    \caption{Normal estimation results comparison using the PGP10 metric (higher is better). }
    \label{appendix:results:pgp10}
\end{table*}

\begin{table*}
    \centering
        \begin{tabular}{|M{0.12\textwidth}| M{0.06\textwidth}|M{0.06\textwidth}|M{0.06\textwidth}|M{0.06\textwidth}|M{0.06\textwidth}|M{0.06\textwidth}|M{0.06\textwidth}|M{0.06\textwidth}|M{0.09\textwidth}|}
    \hline
\textbf{Aug.} & \multicolumn{2}{c|}{\textbf{PCPNet} \cite{guerrero2018pcpnet}}& \multicolumn{3}{c|}{\textbf{Jet} \cite{cazals2005estimating}} & \multicolumn{3}{c|}{\textbf{PCA} \cite{hoppe1992surface}} & \textbf{NestiNet}\\
\hline
Scale&ss&ms&small&med&large&small&med&large&MoE\\
\hline
None &0.7078&0.6986&0.7905&0.6284&0.5395&0.7756&0.6192&0.5361&\textbf{0.8057}\\
\hline
Noise   &&&&&&&&&\\
$\sigma = 0.00125$ &0.6245&0.5932&0.4132&0.6237&0.5377&0.4758&0.6157&0.5335&\textbf{0.6611}\\
$\sigma = 0.006$   &0.4486&0.366&0.027&0.4152&0.4837&0.02998&0.42&0.4812&\textbf{0.5618}\\
$\sigma = 0.01$    &0.3156&0.2482&0.0099&0.1462&0.3715&0.0104&0.154&0.3719&\textbf{0.399}\\
\hline
Density &&&&&&&&&\\
Gradient&0.6065&0.6254&\textbf{0.7883}&0.6442&0.4976&0.7743&0.647&0.4894&0.7749\\
Striped&0.6126&0.6231&\textbf{0.7753}&0.6321&0.4598&0.7575&0.6174&0.4415&0.7676\\
\hline
Average&0.5526&0.5257&0.4674&0.5150&0.4816&0.4706&0.5122&0.4756&\textbf{0.6617}\\
\hline
    \end{tabular}
    \caption{Normal estimation results comparison using the PGP5 metric (higher is better).}
    \label{appendix:results:pgp5}
\end{table*}

\begin{figure}
     \centering
     \includegraphics[width=0.46\textwidth]{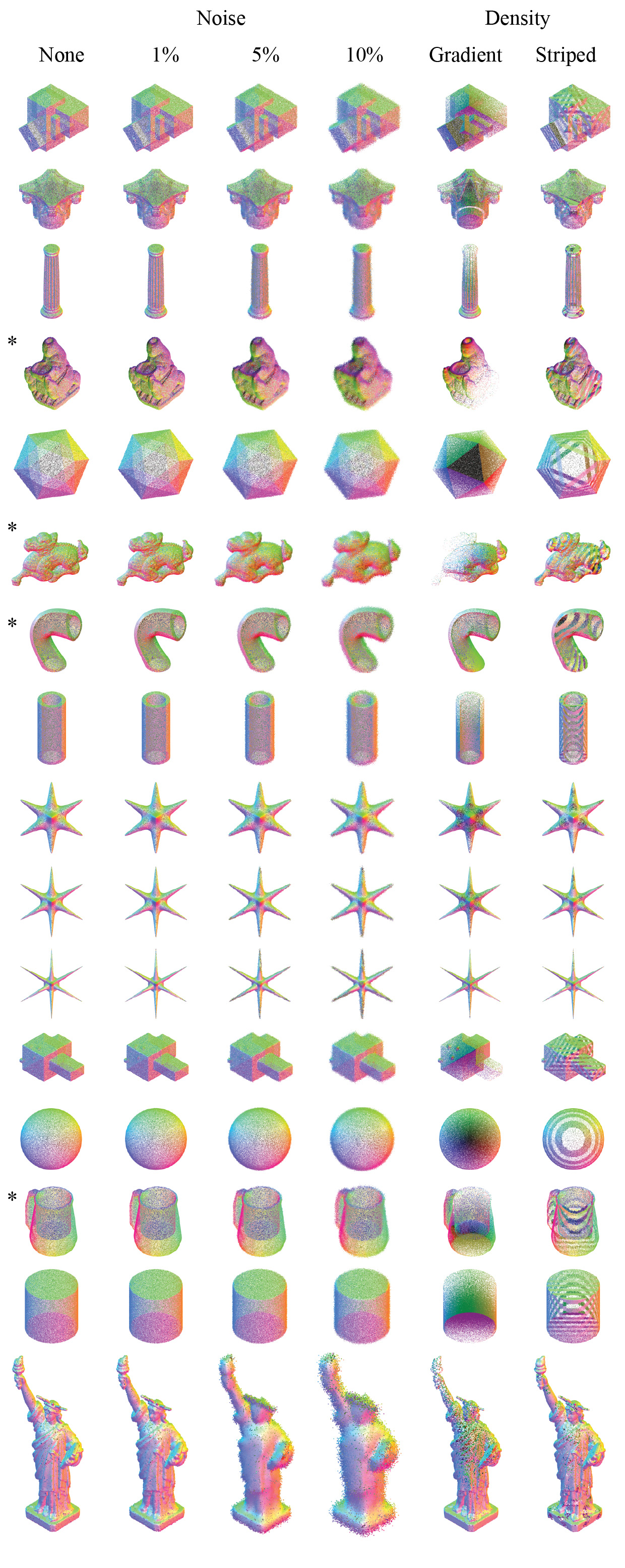}
     \caption{Nesti-Net normal prediction results for different noise levels (columns 1-4) and density distortions (columns 5-6). The point colors are normal vectors mapped to RGB. Point clouds in rows marked with * were rotated for a better view angle.}
     \label{fig:appendix:normal_vis}
 \end{figure}

\begin{figure}
     \centering
     \includegraphics[width=0.46\textwidth]{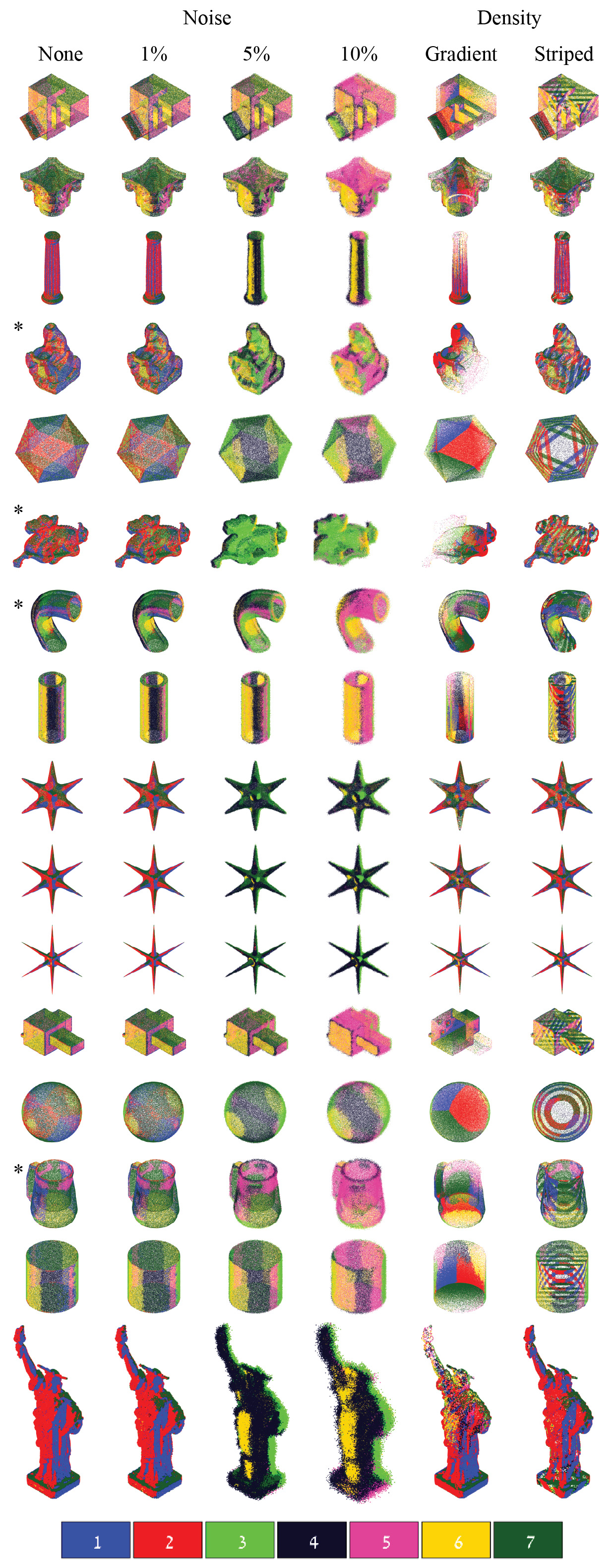}
     \caption{Nesti-Net predicted experts (scales). Each color represents the predicted expert for optimal normal estimation. Color coding is given at the bottom. Point clouds in rows marked with * were rotated for a better view angle.}
     \label{fig:appendix:expert_vis}
 \end{figure}

\begin{figure}
     \centering
     \includegraphics[width=0.46\textwidth]{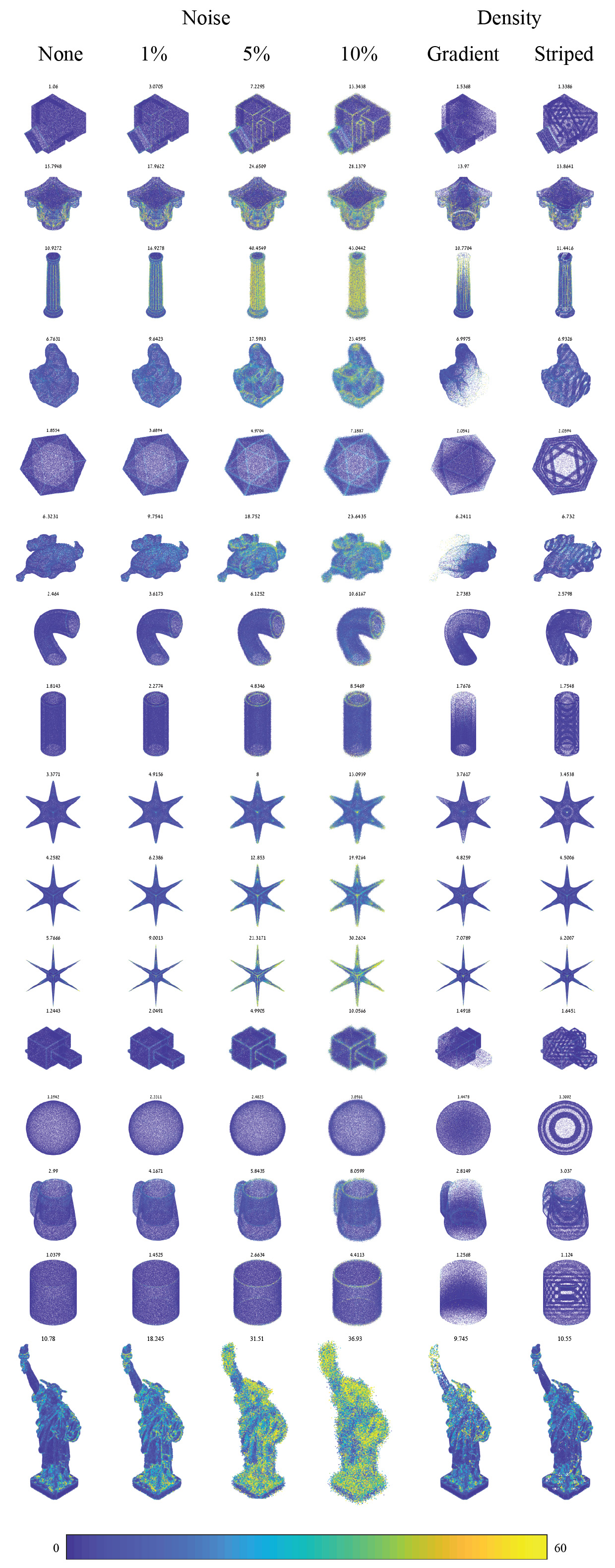}
     \caption{Normal estimation error results for Nesti-Net compared to other methods for different noise levels (columns 1-4) and density distortions (columns 5-6). The point colors correspond to angular difference, mapped to a heatmap between 0-60; see bottom color bar. The number above each point cloud is its RMS error.}
     \label{fig:appendix:error_vis}
 \end{figure}

 We show here additional results from section \ref{SubSec:results:baseline_n_est}. Figure \ref{fig:appendix:normal_vis} shows normal vectors mapped to RGB  color at each point. Figure \ref{fig:appendix:error_vis} shows the angular error mapped to a heatmap between 0-60. The number above each point cloud is its RMS error. Figure \ref{fig:appendix:expert_vis} shows the expert (scale) prediction by assigning a color to each expert and visualizing the chosen expert color over the point cloud.


\begin{figure}
	     \centering
     \includegraphics[width=0.48\textwidth]{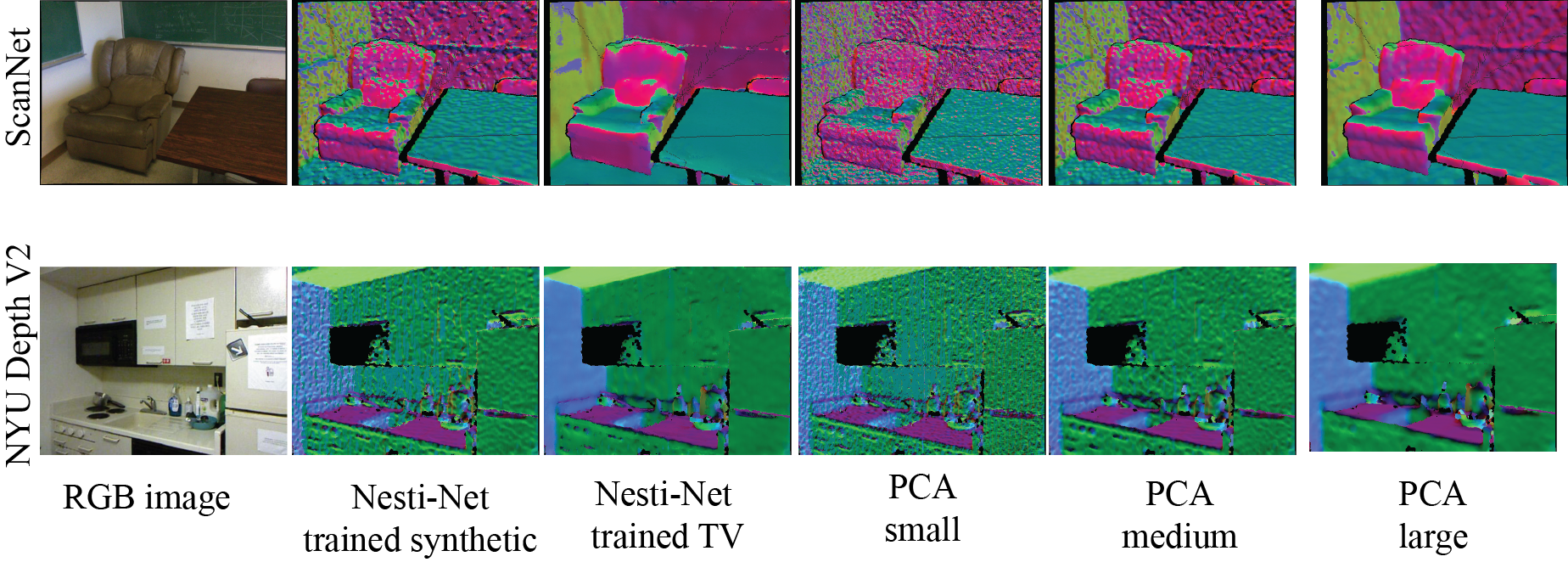}
     \caption{Normal estimation results on scanned data from the NYU Depth V2 \cite{SilbermanECCV12} dataset and the ScanNet \cite{dai2017scannet} dataset. }
     \label{fig:appendix:real_scanned_results}
\end{figure}

In Section \ref{subSec:results:scanned_data} we report the normal estimation of  Nesti-Net and compare it qualitatively to the PCA results with medium scale. For additional comparison, \ref{fig:appendix:real_scanned_results} shows  results of PCA for small and large scale. It shows that a small scale produces a noisy output and a large scale over-smooths fine details and corners. 

\subsection{Time complexity and timing}
\begin{figure}
    \centering
    \includegraphics[width=0.98\linewidth]{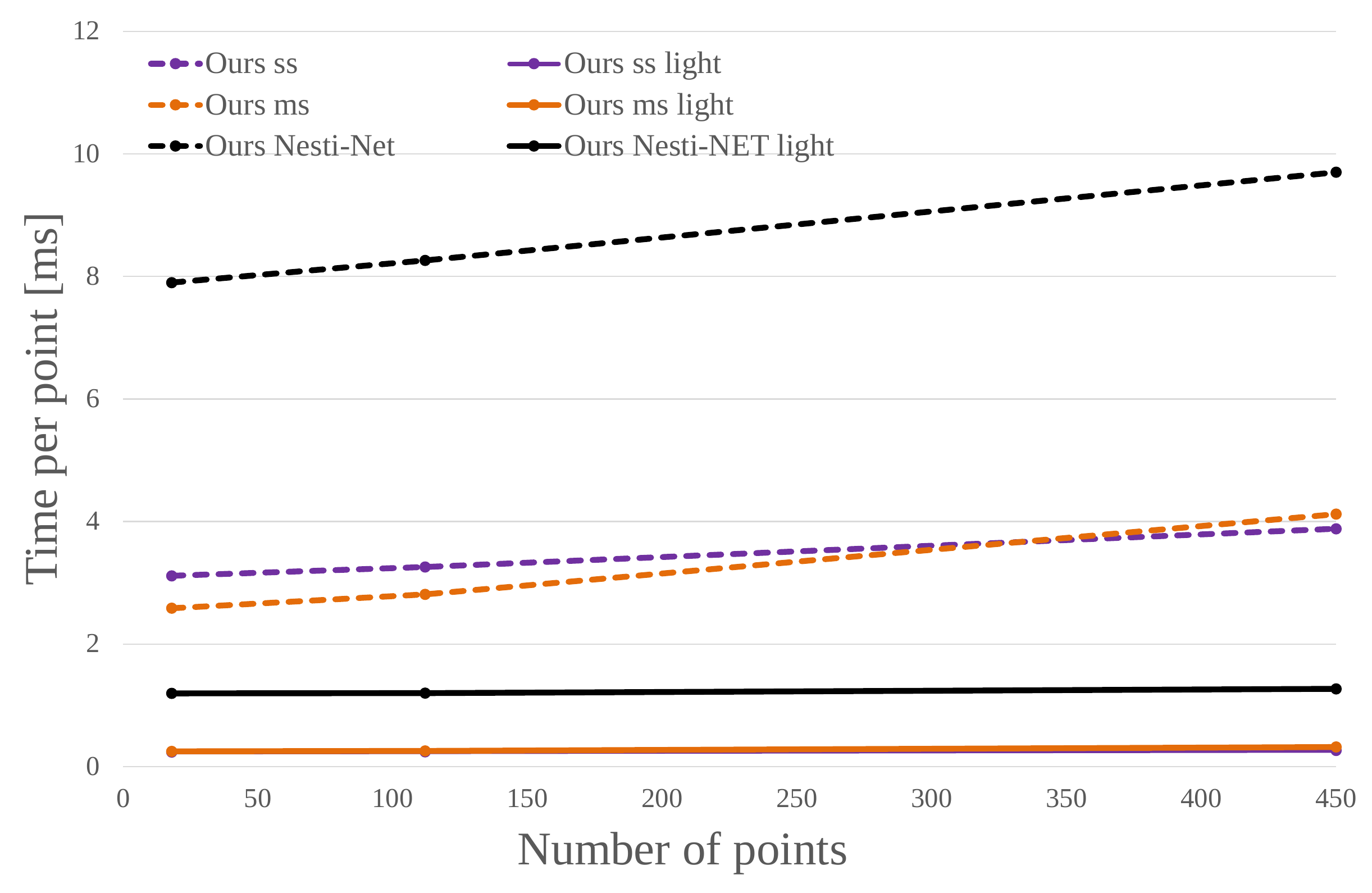}
    \caption{Timing results for our method and its ablations using $8^3$ Gaussians and $3^3$ Gaussians in the 'light' version. Ablations include single-scale (ss), multi-scale (ms) and mixture-of-experts (Nesti-Net). Time is measured in ms per point }
    \label{fig:appendix:timing_ours}
\end{figure}

\begin{figure}
    \centering
    \includegraphics[width=0.98\linewidth]{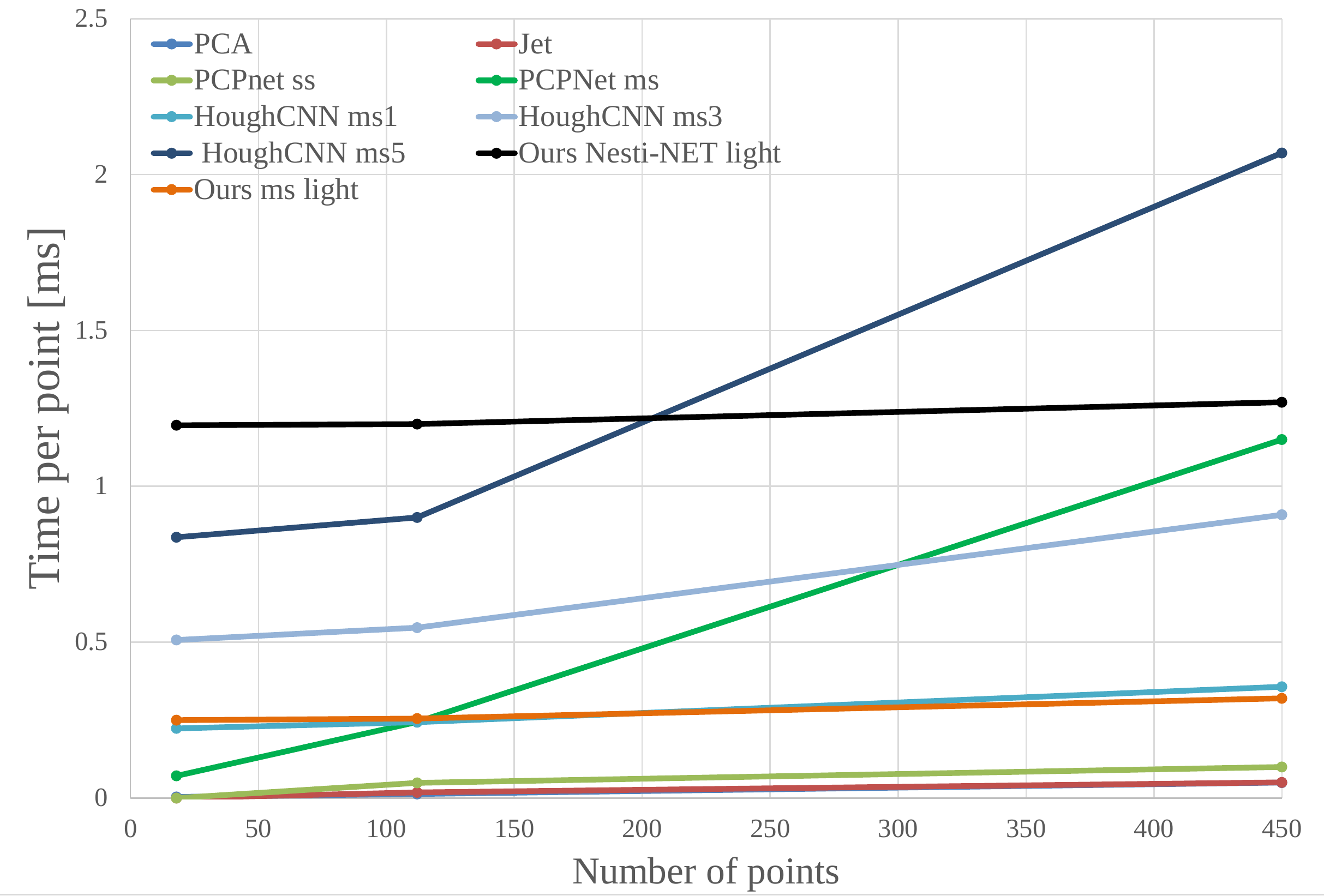}
    \caption{Timing results for normal estimating methods measured in ms per point.}
    \label{fig:appendix:timing_all}
\end{figure}

We subdivide Nesti-Net's time complexity into its two main stages: MuPS computation and normal estimation. It was shown in \cite{ben20183dmfv} that the time complexity of 3DmFV is $O(KT)$. Here $K$ is the number of Gaussians and $T$ is the number of points in the point cloud. MuPS computes 3DmFV of $n$ scales (point subsets) containing a maximum of $T_{max}$ points. Therefore, its time complexity is $O(nKT_{max})$. 
The time complexity of the normal estimation network is constant and proportional to the number of operators in the network. Adding experts to the network increases training time but does not affect test time since only one expert is evaluated during test time. Adding additional scales, however, affects the scale manager network by introducing additional operators. Nevertheless, the normal estimation time is independent of the number of points.
We report the time performance of our method and its ablations in Figure \ref{fig:appendix:timing_ours}. It includes timing results for single-scale (ss), multi-scale (ms) and mixture-of-experts (Nesti-Net) using $8^3$ Gaussians and $3^3$ Gaussians in the 'light' versions. Timing is measured as a function of the number of points within each scale.  Figure \ref{fig:appendix:timing_ours} shows that choosing a lower number of Gaussians for the MuPS representation significantly improves speed but introduces a tradeoff with accuracy. For example, the average RMS error of 'Nesti-Net light' is $13.5$, which is still superior to all other methods but by a smaller margin. 
We also report the timing results of different methods in Figure \ref{fig:appendix:timing_all} and compare to our 'light' version.  Note that the methods were implemented using different frameworks; PCA and Jet were implemented as part of the CGAL library, PCPNet uses pytorch, HoughCNN uses Cuda code directly, and Nesti-Net was implemented using TensorFlow. All measurements were performed on the same machine  with a quad-core Intel i7-4770 CPU, 16GB RAM, and an Nvidia GTX 1080 GPU. The figure shows that PCA and Jet are the fastest methods (PCA is slightly faster) and that the learning-based approaches are comparable. All of the results are outside the range of real-time performance. The figure also shows that our method's timing is not as sensitive to the number of points within each scale as the other methods.

\end{document}